# Logarithmic-Time Updates and Queries in Probabilistic Networks

**Arthur L. Delcher** DELCHER@CS.LOYOLA.EDU
*Computer Science Department, Loyola College in Maryland*
*Baltimore, MD 21210*

**Adam J. Grove** GROVE@RESEARCH.NJ.NEC.COM
*NEC Research Institute*
*Princeton, NJ 08540*

**Simon Kasif** KASIF@CS.JHU.EDU
*Department of Computer Science, Johns Hopkins University*
*Baltimore, MD 21218*

**Judea Pearl** PEARL@LANAI.CS.UCLA.EDU
*Department of Computer Science, University of California*
*Los Angeles, CA 90095*

## Abstract

Traditional databases commonly support efficient query and update procedures that operate in time which is sublinear in the size of the database. Our goal in this paper is to take a first step toward dynamic reasoning in probabilistic databases with comparable efficiency. We propose a dynamic data structure that supports efficient algorithms for updating and querying singly connected Bayesian networks. In the conventional algorithm, new evidence is absorbed in time $O(1)$ and queries are processed in time $O(N)$, where $N$ is the size of the network. We propose an algorithm which, after a preprocessing phase, allows us to answer queries in time $O(\log N)$ at the expense of $O(\log N)$ time per evidence absorption. The usefulness of sub-linear processing time manifests itself in applications requiring (near) real-time response over large probabilistic databases. We briefly discuss a potential application of dynamic probabilistic reasoning in computational biology.

## 1. Introduction

Probabilistic (Bayesian) networks are an increasingly popular modeling technique that has been used successfully in numerous applications of intelligent systems such as real-time planning and navigation, model-based diagnosis, information retrieval, classification, Bayesian forecasting, natural language processing, computer vision, medical informatics and computational biology. Probabilistic networks allow the user to describe the environment using a "probabilistic database" that consists of a large number of random variables, each corresponding to an important parameter in the environment. Some random variables could in fact be hidden and may correspond to some unknown parameters (causes) that influence the observable variables. Probabilistic networks are quite general and can store information such as the probability of failure of a particular component in a computer system, the prob-ability of page $i$ in a computer cache being requested in the near future, the probability of a document being relevant to a particular query, or the probability of an amino-acid subsequence in a protein chain folding into an alpha-helix conformation.

The applications we have in mind include networks that are dynamically maintained to keep track of a *probabilistic* model of a changing system. For instance, consider the task of automated detection of power-plant failures. We might repeat a cycle that consists of the following sequence of operations: First we perform sensing operations. These operations cause updates to be performed to specific variables in the probabilistic database. Based on this evidence we estimate (query) the probability of failure in certain sites. More precisely, we query the probability distribution of the random variables that measure the probability of failure in these sites based on the evidence. Since the plant requires constant monitoring, we must repeat the cycle of sense/evaluate on a frequent basis.

A conventional (non-probabilistic) database tracking the plant's state would not be appropriate here, because it is not possible to directly observe whether a failure is about to occur. On the other hand, a probabilistic "database" based on a Bayesian network will only be useful if the operations—update and query—can be performed very quickly. Because real-time or near real-time is so often necessary, the question of doing *extremely* fast reasoning in probabilistic networks is important.

Traditional (non-probabilistic) databases support efficient query and update procedures that often operate in time which is sublinear in the size of the database (*e.g.*, using binary search). Our goal in this paper is to take a step toward systems that can perform dynamic probabilistic reasoning (such as what is the probability of an event given a set of observations) in time which is sublinear in the size of the probabilistic network. Typically, sublinear performance in complex networks is attained by using parallelism. This paper relies on preprocessing.

Specifically, we describe new algorithms for performing queries and updates in belief networks in the form of trees (causal trees, polytrees and join trees). We define two natural database operations on probabilistic networks.

1. Update-Node: Perform sensory input, modify the evidence at a leaf node (single variable) in the network and **absorb** this evidence into the network.

2. Query-Node: Obtain the marginal probability distribution over the values of an arbitrary node (single variable) in the network.

The standard algorithms introduced by Pearl (1988) can perform the Query-Node operation in $O(1)$ time although evidence absorption, *i.e.*, the Update-Node operation, takes $O(N)$ time where $N$ is the size of the network. Alternatively, one can assume that the Update-Node operation takes O(1) time (by simply recording the change) and the Query-Node operation takes $O(N)$ time (evaluating the entire network).

In this paper we describe an approach to perform both queries and updates in $O(\log N)$ time. This can be very significant in some systems since we improve the ability of a system to respond after a change has been encountered from $O(N)$ time to $O(\log N)$. Our approach is based on preprocessing the network using a form of node absorption in a carefully structured way to create a hierarchy of abstractions of the network. Previous uses of node absorption techniques were reported by Peot and Shachter (1991).

We note that measuring complexity only in terms of the size of the network, $N$, can overlook some important factors. Suppose that each variable in the network has domain size $k$ or less. For many purposes, $k$ can be considered constant. Nevertheless, some of the algorithms we consider have a slowdown which is some power of $k$, which can be become significant in practice unless $N$ is very large. Thus we will be careful to state this slowdown where it exists.

Section 2 considers the case of causal trees, *i.e.*, singly connected networks in which each node has at most one parent. The standard algorithm (see Pearl, 1988) must use $O(k^2N)$ time for either updates or for retrieval, although one of these operations can be done in $O(1)$ time. As we discuss briefly in Section 2.1, there is also a straightforward variant on this algorithm that takes $O(k^2D)$ time for both queries and updates, where $D$ is the height of the tree.

We then present an algorithm that takes $O(k^3 \log N)$ time for updates and $O(k^2 \log N)$ time for queries in any causal tree. This can of course represent a tremendous speedup, especially for large networks. Our algorithm begins with a polynomial-time preprocessing step (linear in the size of the network), constructing another data structure (which is not itself a probabilistic tree) that supports fast queries and updates. The techniques we use are motivated by earlier algorithms for dynamic arithmetic trees, and involve "caching" sufficient intermediate computations during the update phase so that querying is also relatively easy. We note, however, that there are substantial and interesting differences between the algorithm for probabilistic networks and those for arithmetic trees. In particular, as will be apparent later, computation in probabilistic trees requires both bottom-up and top-down processing, whereas arithmetic trees need only the former. Perhaps even more interesting is that the relevant probabilistic operations have a different algebraic structure than arithmetic operations (for instance, they lack distributivity).

Bayesian trees have many applications in the literature including classification. For instance, one of the most popular methods for classification is the Bayes classifier that makes independence assumption on the features that are used to perform classification (Duda & Hart, 1973; Rachlin, Kasif, Salzberg, & Aha, 1994). Probabilistic trees have been used in computer vision (Hel-Or & Werman, 1992; Chelberg, 1990), signal processing (Wilsky, 1993), game playing (Delcher & Kasif, 1992), and statistical mechanics (Berger & Ye, 1990). Nevertheless, causal trees are fairly limited for modeling purposes. However similar structures, called *join trees*, arise in the course of one of the standard algorithms for computing with arbitrary Bayesian networks (see Lauritzen and Spiegelhalter, 1988). Thus our algorithm for join trees has potential relevance to many networks that are not trees. Because join trees have some special structure, they allow some optimization of the basic causal-tree algorithm. We elaborate on this in Section 5.

In Section 6 we consider the case of arbitrary polytrees. We give an $O(\log N)$ algorithm for updates and queries, which involves transforming the polytree to a join tree, and then using the results of Sections 2 and 5. The join tree of a polytree has a particularly simple form, giving an algorithm in which updates take $O(k^{p+3} \log N)$ time and queries $O(k^{p+2} \log N)$, where $p$ is the maximum number of parents of any node. Although the constant appears large, it must be noted that the original polytree takes $O(k^{p+1}N)$ space merely to represent, if conditional probability tables are given as explicit matrices.

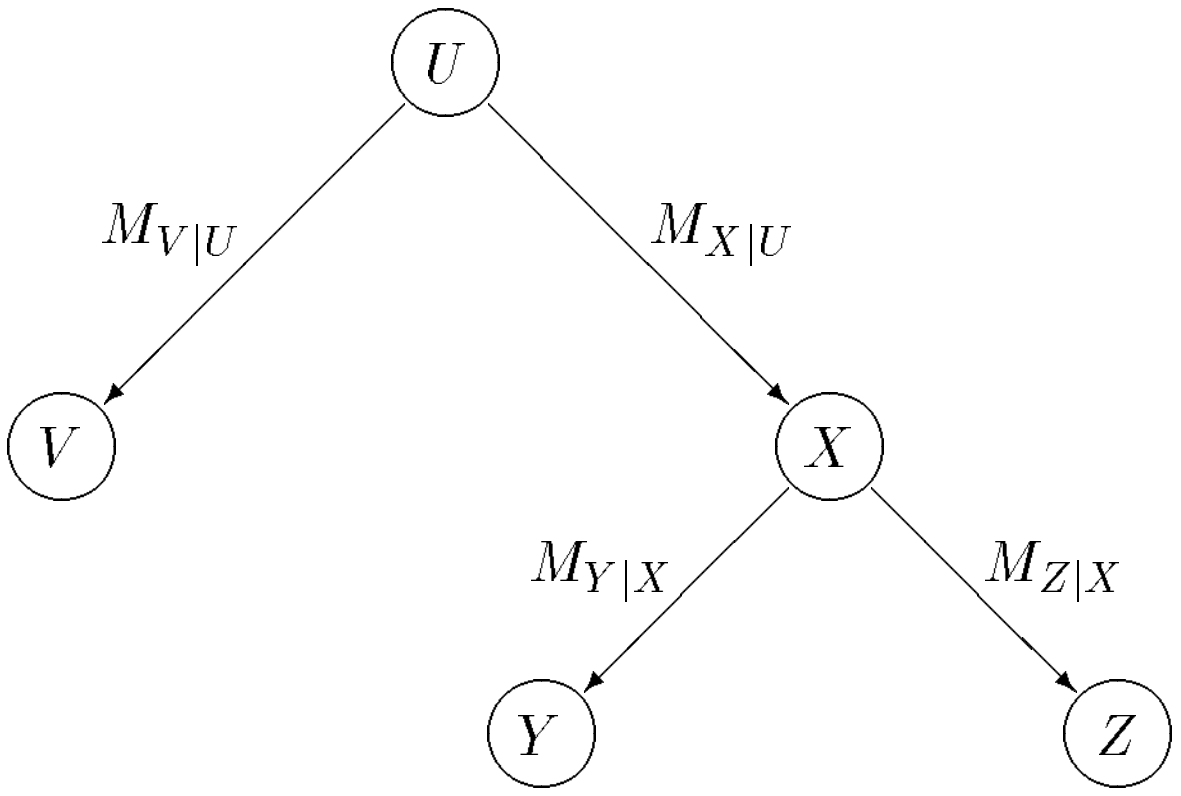

Figure 1: A segment of a causal tree.

Finally, we discuss a specific modelling application in computational biology where probabilistic models are used to describe, analyze and predict the functional behavior of biological sequences such as protein chains or DNA sequences (see Delcher, Kasif, Goldberg, and Hsu, 1993 for references). Much of the information in computational biology databases is noisy. However, a number of successful attempts to build probabilistic models have been made. In this case, we use a probabilistic tree of depth 300 that consists of 600 nodes and all the matrices of conditional probabilities are $2 \times 2$. The tree is used to model the dependence of a protein's secondary structure on its chemical structure. The detailed description of the problem and experimental results are given by Delcher et al. (1993). For this problem we obtain an effective speed-up of about a factor of 10 to perform an update as compared to the standard algorithm. Clearly, getting an order of magnitude improvement in the response time of a probabilistic real-time system could be of tremendous importance in future use of such systems.

## 2. Causal Trees

A probabilistic causal tree is a directed tree in which each node represents a discrete random variable $X$, and each directed edge is annotated by a matrix of conditional probabilities $M_{Y|X}$ (associated with edge $X \rightarrow Y$). That is, if $x$ is a possible value of $X$, and $y$ of $Y$, then the $(x,y)$th component of $M_{Y|X}$ is $Pr(Y = y|X = x)$. Such a tree represents a joint probability distribution over the product space of all variables; for detailed definitions and discussion see Pearl (1988). Briefly, the idea is that we consider the product, over all nodes, of the conditional probability of the node given its parents. For example, in Figure 1 the implied distribution is:

$$Pr(U = u, V = v, X = x, Y = y, Z = z) = \\ Pr(U = u)\, Pr(V = v|U = u)\, Pr(X = x|U = u)\, Pr(Y = y|X = x)\, Pr(Z = z|X = x).$$

Given particular values of $u, v, x, y, z$, the conditional probabilities can be read from the appropriate matrices $M$. One advantage of such a product representation is that it is very concise. In this example, we need four matrices and the unconditional probability over $U$, but the size of each is at most the square of the largest variable's domain size. In contrast, a general distribution over $N$ variables requires an exponential (in $N$) representation.

Of course, not every distribution can be represented as a causal tree. But it turns out that the product decomposition implied by the tree corresponds to a particular pattern of conditional independencies which often hold (if perhaps only approximately) in real applications. Intuitively speaking, in Figure 1 some of these implied independencies are that the conditional probability of $U$ given $V$, $X$, $Y$ and $Z$ depends only on values of $V$ and $X$; and the probability of $Y$ given $U$, $V$, $X$, and $Z$ depends only on $X$. Independencies of this sort can arise for many reasons, for instance from a causal modeling of the interactions between the variables. We refer the reader to Pearl (1988) for details related to the modeling of independence assumptions using graphs.

In the following, we make several assumptions that significantly simplify the presentation, but do not sacrifice generality. First, we assume that each variable ranges over the same, constant, number of values $k$.[1] It follows that the marginal probability distribution for each variable can be viewed as a $k$-dimensional vector, and each conditional probability matrix such as $M_{Y|X}$ is a square $k \times k$ matrix. A common case is that of binary random variables ($k = 2$); the distribution over the values (TRUE, FALSE) is then $(p, 1-p)$ for some probability $p$.

The next assumption is that the tree is binary, and complete, so that each node has 0 or 2 children. Any tree can be converted into this form, by at most doubling the number of nodes. For instance, suppose node $p$ has children $c_1, c_2, c_3$ in the original tree. We can create another "copy" of $p$, $p'$, and rearrange the tree such that the two children of $p$ are $c_1$ and $p'$, and the two children of $p'$ are $c_2$ and $c_3$. We can constrain $p'$ always to have the same value as $p$ simply by choosing the identity matrix for the conditional probability table between $p$ and $p'$. Then the distribution represented by the new tree is effectively the same as the original. Similarly, we can always add "dummy" leaf nodes if necessary to ensure a node has two children. As explained in the introduction, we are interested in processes in which certain variables' values are observed, upon which we wish to condition. Our final assumption is that these observed **evidence** nodes are all leaves of the tree. Again, because it is possible to "copy" nodes and to add dummy nodes, this is not restrictive.

The product distribution alluded to above corresponds to the distribution over variables prior to any observations. In practice, we are more interested in the conditional distribution, which is simply the result of conditioning on all the observed evidence (which, by the earlier assumption, corresponds to seeing values for all the leaf nodes). Thus, for each non-leaf node $X$ we are interested in the conditional marginal probability over $X$, *i.e.*, the $k$-dimensional vector:

$$Bel(X) = Pr(X \mid \text{all evidence values}).$$

The main algorithmic problem is to compute $Bel(X)$ for each (non-evidence) node $X$ in the tree given the current evidence. It is well known that the probability vector $Bel(X)$ can be computed in linear time (in the size of the tree) by a popular algorithm based on

1. This assumption is nonrestrictive because we can add "dummy" values to each variable's range, which should be given conditional probability 0. Nevertheless, there may some computational advantage in allowing different variable domain sizes. The changes required to permit this are not difficult, but since they complicate the presentation somewhat we omit them.

the following equation:

$$Bel(X) = Pr(X \mid \text{all evidence}) = \alpha * \lambda(X) * \pi(X)$$

Here $\alpha$ is a normalizing constant, $\lambda(X)$ is the probability of all the evidence in the subtree below node $X$ given $X$, and $\pi(X)$ is the probability of $X$ given all evidence in the rest of the tree. To interpret this equation, note that if $X = (x_1, x_2, \ldots, x_k)$ and $(Y = y_1, y_2, \ldots, y_k)$ are two vectors we define $*$ to be the operation of component-wise product (pairwise or dyadic product of vectors):

$$X * Y = (x_1 y_1, x_2 y_2, \ldots, x_k y_k).$$

The usefulness of $\lambda(X)$ and $\pi(X)$ derives from the fact that they can be computed recursively, as follows:

1. If $X$ is the root node, $\pi(X)$ is the prior probability of $X$.
2. If $X$ is a leaf node, $\lambda(X)$ is a vector with 1 in the $i$th position (where the $i$th value has been observed) and 0 elsewhere. If no value for $X$ has been observed, then $\lambda(X)$ is a vector consisting of all 1's.[2]
3. Otherwise, if, as shown in Figure 1, the children of node $X$ are $Y$ and $Z$, its sibling is $V$ and its parent is $U$, we have:

$$\begin{aligned} \lambda(X) &= (M_{Y|X} \cdot \lambda(Y)) * (M_{Z|X} \cdot \lambda(Z)) \\ \pi(X) &= M^{\mathrm{T}}_{X|U} \cdot \Big(\pi(U) * (M_{V|U} \cdot \lambda(V))\Big) \end{aligned}$$

Our presentation of this technique follows that of Pearl (1988). However, we use a somewhat different notation in that we don't describe messages sent to parents or successors, but rather discuss the direct relations among the $\pi$ and $\lambda$ vectors in terms of simple algebraic equations. We will take advantage of algebraic properties of these equations in our development.

It is very easy to see that the equations above can be evaluated in time proportional to the size of the network. The formal proof is given by Pearl (1988).

**Theorem 1:** The belief distribution of every variable (that is, the marginal probability distribution for each variable, given the evidence) in a causal tree can be evaluated in $O(k^2 N)$ time where $N$ is the size of the tree. (The factor $k^2$ is due to the multiplication of a matrix by a vector that must be performed at each node.)

This theorem shows that it is possible to perform evidence absorption in $O(N)$ time, and queries in constant time (*i.e.*, by retrieving the previously computed values from a lookup table). In the next sections we will show how to perform both queries and updates in worst-case $O(\log N)$ time. Intuitively, we will not recompute all the marginal distributions after an update, but rather make only a small number of changes, sufficient, however, to compute the value of any variable with only a logarithmic delay.

2. Or we can set to 1 all components corresponding to *possible* values—this is especially useful when the observed variable is part of a joint-tree clique (Section 5). In general, $\lambda(X)$ should be thought of as the *likelihood* vector over $X$ given our observations about $X$.

### 2.1 A Simple Preprocessing Approach

To obtain intuition about the new approach we begin with a very simple observation. Consider a causal tree $T$ of depth $D$. For each node $X$ in the tree we initially compute its $\lambda(X)$ vector. $\pi$ vectors are left uncomputed. Given an update to a node $Y$, we calculate the revised $\lambda(X)$ vectors for all nodes $X$ that are ancestors of $Y$ in the tree. This clearly can be done in time proportional to the depth of the tree, *i.e.*, $O(D)$. The rest of the information in the tree remains unchanged. Now consider a Query-Node operation for some node $V$ in the tree. We obviously already have the accurate $\lambda(V)$ vector for every node in the tree including $V$. However, in order to compute its $\pi(V)$ vector we need to compute only the $\pi(Y)$ vectors for all the nodes above $V$ in the tree and multiply these by the appropriate $\lambda$ vectors that are kept current. This means that to compute the accurate $\pi(V)$ vector we need to perform $O(D)$ work as well. Thus, in this approach we don't perform the complete update to every $\lambda(X)$ and $\pi(X)$ vector in the tree.

**Lemma 2:** Update-Node and Query-Node operations in a causal tree $T$ can be performed in $O(k^2 D)$ time where $D$ is the depth of the tree.

This implies that if the tree is balanced, both operations can be done in $O(\log N)$ time. However, in some important applications the trees are not balanced (*e.g.*, models of temporal sequences, Delcher et al., 1993). The obvious question therefore is: Given a causal tree $T$ can we produce an equivalent balanced tree $T'$? While the answer to this question appears to be difficult, it is possible to use a more sophisticated approach to produce a data structure (which is *not* a causal tree) to process queries and updates in $O(\log N)$ time. This approach is described in the subsequent sections.

### 2.2 A Dynamic Data Structure For Causal Trees

The data structure that will allow efficient incremental processing of a probabilistic tree $T = T_0$ will be a sequence of trees, $T_0, T_1, T_2, \ldots, T_i, \ldots, T_{\log N}$. Each $T_{i+1}$ will be a *contracted* version of $T_i$, whose nodes are a subset of those in $T_i$. In particular, $T_{i+1}$ will contain about half as many leaves as its predecessor.

We defer the details of this contraction process until the next section. However, one key idea is that we maintain consistency, in the sense that $Bel(X)$, $\lambda(X)$, and $\pi(X)$ are given the same values by all the trees in which $X$ appears. We choose the conditional probability matrices in the contracted trees (*i.e.*, all trees other than $T_0$) to ensure this.

Recall that the $\lambda$ and $\pi$ equations have the form

$$\begin{aligned}
\lambda(X) &= (M_{Y|X} \cdot \lambda(Y)) * (M_{Z|X} \cdot \lambda(Z)) \\
\pi(X) &= M^{\mathrm{T}}_{X|U} \cdot \Big(\pi(U) * (M_{V|U} \cdot \lambda(V))\Big)
\end{aligned}$$

if $Y$ and $Z$ are children of $X$, $X$ is a right child of $U$, and $V$ is $X$'s sibling (Figure 1). However, these equations are not in the most convenient form and the following notational conventions will be very helpful. First, let $A_i(x)$ (resp., $B_i(x)$) denote the conditional probability matrix between $X$ and $X$'s left (resp., right) child in the tree $T_i$. Note that the identity of these children can differ from tree to tree, because some of $X$'s original children might be removed by the contraction process. One advantage of the new notation is that the explicit dependence on the identity of the children is suppressed. Next, suppose $X$'s parent in $T_i$ is $u$. Then we let $C_i(x)$ denote either $A_i(u)$ or $B_i(u)$, and $D_i(x)$ denote either $B_i(u)^{\mathrm{T}}$ or $A_i(u)^{\mathrm{T}}$, depending on whether $X$ is the right or left child, respectively, of $U$. It will not be necessary to keep careful track of these correspondences, but simply to note that the above equations become:[3]

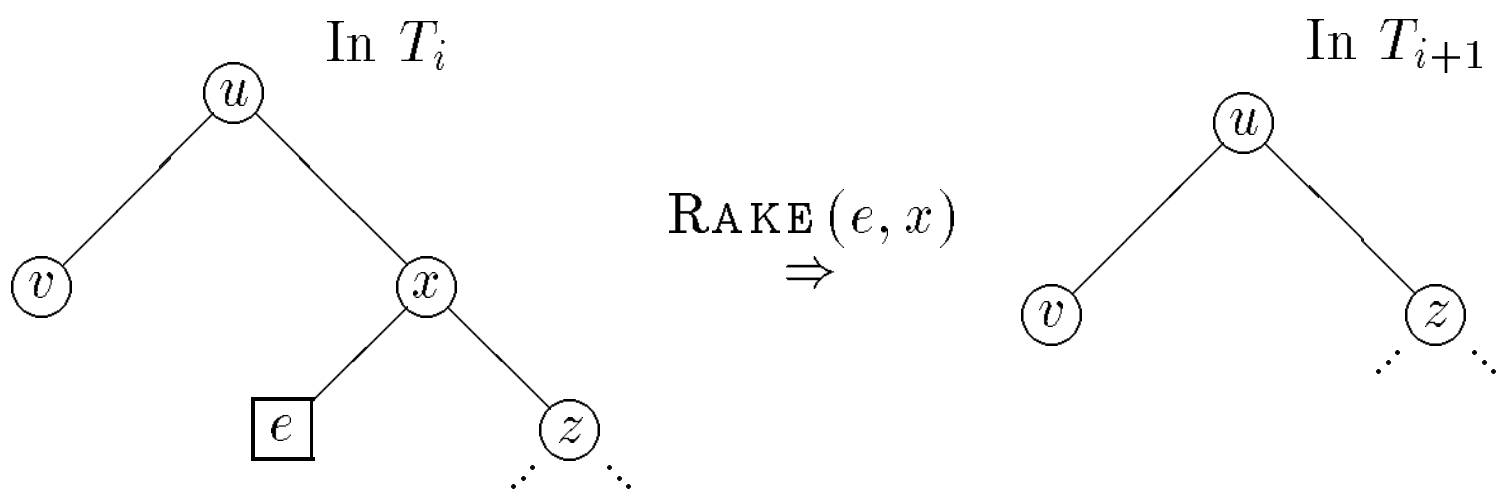

Figure 2: The effect of the operation RAKE $(e, x)$. $e$ must be a leaf, but $z$ may or may not be a leaf.

$$\begin{aligned} \lambda(x) &= A_i(x) \cdot \lambda(y) * B_i(x) \cdot \lambda(z) \\ \pi(x) &= D_i(x) \cdot (\pi(u) * C_i(x) \cdot \lambda(v)) \end{aligned}$$

In the next section we describe the preprocessing step that creates the dynamic data structure.

### 2.3 RAKE Operation

The basic operation used to contract the tree is RAKE which removes both a leaf and its parent from the tree. The effect of this operation on the tree is shown in Figure 2. We now define the algebraic effect of this operation on the equations associated with this tree. Recall that we want to define the conditional probability matrices in the raked tree so that the distribution over the remaining variables is unchanged. We achieve this by substituting the equations for $\lambda(x)$ and $\pi(x)$ into the equations for $\lambda(u)$, $\pi(z)$, and $\pi(v)$. In the following, it is important to note that $\pi(u)$, $\lambda(z)$ and $\lambda(v)$ are unaffected by the rake operation.

In the following, let $\mathrm{Diag}_\alpha$ denote the diagonal matrix whose diagonal entries are the components of the vector $\alpha$. We derive the algebraic effect of the rake operation as follows:

$$\begin{aligned} \lambda(u) &= A_i(u) \cdot \lambda(v) * B_i(u) \cdot \lambda(x) \\ &= A_i(u) \cdot \lambda(v) * B_i(u) \cdot (A_i(x) \cdot \lambda(e) * B_i(x) \cdot \lambda(z)) \\ &= A_i(u) \cdot \lambda(v) * B_i(u) \cdot \left(\mathrm{Diag}_{A_i(x) \cdot \lambda(e)} \cdot B_i(x) \cdot \lambda(z)\right) \\ &= A_i(u) \cdot \lambda(v) * \left(B_i(u) \cdot \mathrm{Diag}_{A_i(x) \cdot \lambda(e)} \cdot B_i(x)\right) \cdot \lambda(z) \\ &= A_{i+1}(u) \cdot \lambda(v) * B_{i+1}(u) \cdot \lambda(z) \end{aligned}$$

where $A_{i+1}(u) = A_i(u)$ and $B_{i+1}(u) = B_i(u) \cdot \mathrm{Diag}_{A_i(x) \cdot \lambda(e)} \cdot B_i(x)$. (Of course, the case where the leaf being raked is a right child generates analogous equations.) Thus, by defining

3. Throughout, we assume that $*$ has lower precedence than matrix multiplication (indicated by $\cdot$).

$A_{i+1}(u)$ and $B_{i+1}(u)$ in this way, we ensure that all $\lambda$ values in the raked tree are identical to the corresponding values in the original tree. This is not yet enough, because we must check that $\pi$ values are similarly preserved. The only two values that could possibly change are $\pi(z)$ and $\pi(v)$, so we check them both. For the former, we must have

$$\begin{aligned} \pi(z) &= D_i(z) \cdot (\pi(x) * C_i(z) \cdot \lambda(e)) \\ &= D_{i+1}(z) \cdot (\pi(u) * C_{i+1}(z) \cdot \lambda(v)). \end{aligned}$$

After substituting for $\pi(x)$ and some algebraic manipulation, we see that this is assured if $C_{i+1}(z) = C_i(x)$ and $D_{i+1}(z) = D_i(z) \cdot \mathrm{Diag}_{C_i(z)\cdot\lambda(e)} \cdot D_i(x)$. However recall that, by definition, $C_{i+1}(z) = A_{i+1}(u)$ and $C_i(x) = A_i(u)$, and so $C_{i+1}(z) = C_i(x)$ follows. Furthermore,

$$\begin{aligned} D_{i+1}(z) &= B_{i+1}(u)^{\mathrm{T}} \\ &= (B_i(u) \cdot \mathrm{Diag}_{A_i(x)\cdot\lambda(e)} \cdot B_i(x))^{\mathrm{T}} \\ &= B_i(x)^{\mathrm{T}} \cdot \mathrm{Diag}_{A_i(x)\cdot\lambda(e)} \cdot B_i(u)^{\mathrm{T}} \\ &= D_i(z) \cdot \mathrm{Diag}_{C_i(z)\cdot\lambda(e)} \cdot D_i(x) \end{aligned}$$

as required.
For $\pi(v)$ it is necessary to verify that

$$\begin{aligned} \pi(v) &= D_i(v) \cdot (\pi(u) * C_i(v) \cdot \lambda(x)) \\ &= D_{i+1}(v) \cdot (\pi(u) * C_{i+1}(v) \cdot \lambda(z)). \end{aligned}$$

By substituting for $\lambda(x)$, this can be shown to be true if $D_{i+1}(v) = D_i(v) = A_i(u)^{\mathrm{T}} = A_{i+1}(u)^{\mathrm{T}}$ and $C_{i+1}(v) = C_i(v) \cdot \mathrm{Diag}_{A_i(x)\cdot\lambda(e)} \cdot B_i(x) = B_{i+1}(u)$. But these identities follow by definition, so we are done.

Beginning with the given tree $T = T_0$, each successive tree is constructed by performing a sequence of rakes, so as to rake away about half of the remaining evidence nodes. More specifically, let Contract be the operation in which we apply the Rake operation to every other leaf of a causal tree, in left-to-right order, excluding the leftmost and the rightmost leaf. Let $\{T_i\}$ be the set of causal trees constructed so that $T_{i+1}$ is the causal tree generated from $T_i$ by a single application of Contract. The following result is proved using an easy inductive argument:

**Theorem 3:** Let $T_0$ be a causal tree of size $N$. Then the number of leaves in $T_{i+1}$ is equal to half the leaves in $T_i$ (not counting the two extreme leaves) so that starting with $T_0$, after $O(\log N)$ applications of Contract, we produce a three-node tree: the root, the leftmost leaf and the rightmost leaf.

Below are a few observations about this process:

1. The complexity of Contract is linear in the size of the tree. Additionally, $\log N$ applications of Contract reduce the set of tree equations to a single equation involving the root in $O(N)$ total time.

2. The total space to store all the sets of equations associated with $\{T_i\}_{0 \le i \le \log N}$ is about twice the space required to store the equations for $T_0$.

3. With each equation in $T_{i+1}$ we also store equations that describe the relationship between the conditional probability matrices in $T_{i+1}$ to the matrices in $T_i$. Notice that, even though $T_{i+1}$ is produced from $T_i$ by a series of rake operations, each matrix in $T_{i+1}$ depends directly on matrices present in $T_i$. This would not be the case if we attempted to simultaneously rake adjacent children.

   We regard these equations as part of $T_{i+1}$. So, formally speaking $\{T_i\}$ are causal trees augmented with some auxiliary equations. Each of the contracted trees describes a probability distribution on a subset of the first set of variables that is consistent with the original distribution.

We note that the ideas behind the Rake operation were originally developed by Miller and Reif (1985) in the context of parallel computation of bottom-up arithmetic expression trees (Kosaraju & Delcher, 1988; Karp & Ramachandran, 1990). In contrast, we are using it in the context of incremental update and query operations in sequential computing. A similar data structure to ours was independently proposed by Frederickson (1993) in the context of dynamic arithmetic expression trees, and a different approach for incremental computing on arithmetic trees was developed by Cohen and Tamassia (1991). There are important and interesting differences between the arithmetic expression-tree case and our own. For arithmetic expressions all computation is done bottom-up. However, in probabilistic networks $\pi$-messages must be passed top-down. Furthermore, in arithmetic expressions when two algebraic operations are allowed, we typically require the distributivity of one operation over the other, but the analogous property does not hold for us. In these respects our approach is a substantial generalization of the previous work, while remaining conceptually simple and practical.

## 3. Example: A Chain

To obtain an intuition about the algorithms, we sketch how to generate and utilize the $T_i$, $0 \le i \le \log N$ and their equations to perform $\lambda$-value queries and updates in $O(\log N)$ time on an $N = 2L + 1$ node chain of length $L$. Consider the chain of length 4 in Figure 3, and the trees that are generated by repeated application of Contract to the chain.

The equations that correspond to the contracted trees in the figure are as follows (ignoring trivial equations). Recall that $A_i(x_j)$ is the matrix associated with the left edge of random variable $x_j$ in $T_i$.

$$\left.\begin{array}{rcl}
\lambda(x_1) & = & A_0(x_1) \cdot \lambda(e_1) * B_0(x_1) \cdot \lambda(x_2) \\
\lambda(x_2) & = & A_0(x_2) \cdot \lambda(e_2) * B_0(x_2) \cdot \lambda(x_3) \\
\lambda(x_3) & = & A_0(x_3) \cdot \lambda(e_3) * B_0(x_3) \cdot \lambda(x_4) \\
\lambda(x_4) & = & A_0(x_4) \cdot \lambda(e_4) * B_0(x_4) \cdot \lambda(e_5)
\end{array}\right\} \text{ for } T_0$$

$$\left.\begin{array}{rcl}
\lambda(x_1) & = & A_1(x_1) \cdot \lambda(e_1) * B_1(x_1) \cdot \lambda(x_3) \\
\lambda(x_3) & = & A_1(x_3) \cdot \lambda(e_3) * B_1(x_3) \cdot \lambda(e_5) \\
\text{where} & & \\
B_1(x_1) & = & B_0(x_1) \cdot \text{Diag}_{A_0(x_2)\cdot\lambda(e_2)} \cdot B_0(x_2) \\
B_1(x_3) & = & B_0(x_3) \cdot \text{Diag}_{A_0(x_4)\cdot\lambda(e_4)} \cdot B_0(x_4)
\end{array}\right\} \text{ for } T_1$$

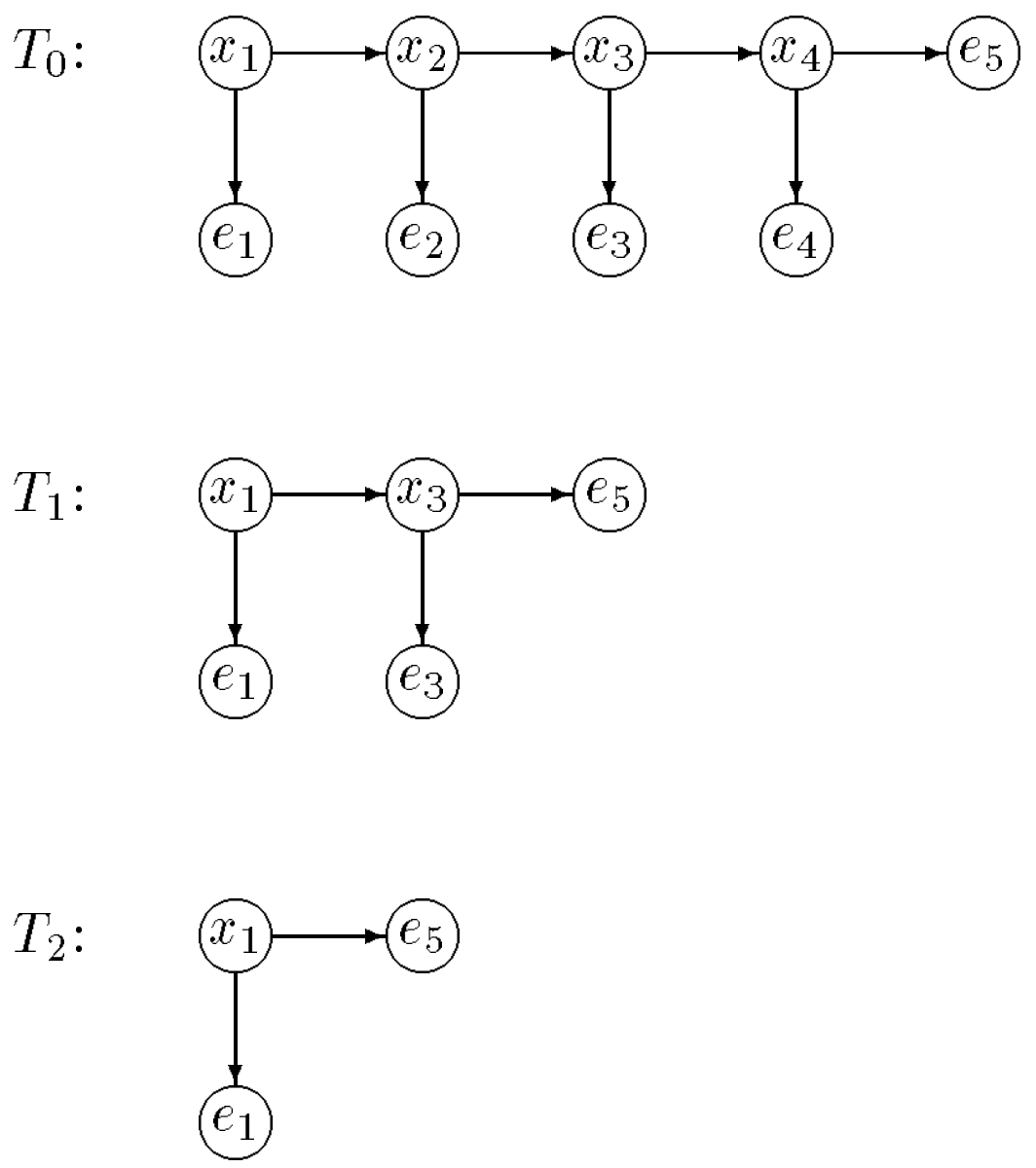

Figure 3: A simple chain example.

$$\left.\begin{array}{rcl} \lambda(x_1) & = & A_2(x_1) \cdot \lambda(e_1) * B_2(x_1) \cdot \lambda(e_5) \\ \text{where} & & \\ B_2(x_1) & = & B_1(x_1) \cdot \text{Diag}_{A_1(x_3) \cdot \lambda(e_3)} \cdot B_1(x_3) \end{array}\right\} \text{ for } T_2$$

We have not listed the $A$ matrices because, in this example, they are constant. Now consider a query operation on $x_2$. Rather than performing the standard computation we will find the level where $x_2$ was "raked". Since this occurred on level 0, we obtain the equation

$$\lambda(x_2) \quad = \quad A_0(x_2) \cdot \lambda(e_2) * B_0(x_2) \cdot \lambda(x_3)$$

Thus we must compute $\lambda(x_3)$, and to do this we find where $x_3$ is "raked". That happened on level 1. However, on that level the equation associated with $x_3$ is:

$$\lambda(x_3) \quad = \quad A_1(x_3) \cdot \lambda(e_3) * B_1(x_3) \cdot \lambda(e_5)$$

That means that we need not follow down the chain. In general for a chain of $N$ nodes we can answer any query to a node on the chain by evaluating $\log N$ equations instead of $N$ equations.

Now consider an update for $e_4$. Since $e_4$ was raked immediately, we first modify the equation

$$B_1(x_3) = B_0(x_3) \cdot \text{Diag}_{A_0(x_4) \cdot \lambda(e_4)} \cdot B_0(x_4)$$

on the first level where $e_4$ occurs on the right-hand side. Since $B_1(x_3)$ is affected by the change to $e_4$, we subsequently modify the equation

$$B_2(x_1) = B_1(x_1) \cdot \text{Diag}_{A_1(x_3) \cdot \lambda(e_3)} \cdot B_1(x_3)$$

on the second level. In general, we clearly need to update at most $\log N$ equations; *i.e.*, one per level. We now generalize this example and describe general algorithms for queries and updates in causal trees.

### 3.1 Performing Queries And Updates Efficiently

In this section we shall show how to utilize the contracted trees $T_i$, $0 \leq i \leq \log N$ to perform queries and updates in $O(\log N)$ time in general causal trees. We shall show that a logarithmic amount of work will be necessary and sufficient to compute enough information in our data structure to update and query any $\lambda$ or $\pi$ value.

### 3.2 $\lambda$ Queries

To compute $\lambda(x)$ for some node $x$ we can do the following. We first locate $ind(x)$, which is defined to be the highest level $i$ such that $x$ appears in $T_i$. The equation for $\lambda(x)$ is of the form:

$$\lambda(x) \quad = \quad A_i(x) \cdot \lambda(y) * B_i(x) \cdot \lambda(z)$$

where $y$ and $z$ are the left and right children, respectively, of $x$ in $T_i$.

Since $x$ does not appear in $T_{i+1}$, it was raked at this level of equations, which implies that one child (we assume $z$) is a leaf. We therefore only need to compute $\lambda(y)$, which can be done recursively. If instead $y$ was the raked leaf, we would compute $\lambda(z)$ recursively.

In either case $O(1)$ operations are done in addition to one recursive call, which is to a value at a higher level of equations. Since there are $O(\log N)$ levels, and the only operations are matrix by vector multiplications, the procedure takes $O(k^2 \log N)$ time. The function $\lambda$-Query $(x)$ is given in Figure 4.

### 3.3 Updates

We now describe how the update operations can modify enough information in the data structure to allow us to query the $\lambda$ vectors and $\pi$ vectors efficiently. Most importantly the reader should note that the update operation does not try to maintain the correct $\pi$ and $\lambda$ values. It is sufficient to ensure that, for all $i$ and $x$, the matrices $A_i(x)$ and $B_i(x)$ (and thus also $C_i(x)$ and $D_i(x)$) are always up to date.

When we update the value of an evidence node, we are simply changing the $\lambda$ value of some leaf $e$. At each level of equations, the value of $\lambda(e)$ can appear at most twice: once in the $\lambda$-equation of $e$'s parent and once in the $\pi$-equation of $e$'s sibling in $T_i$. When $e$ disappears, say at level $i$, its value is incorporated into one of the constant matrices $A_{i+1}(u)$ or $B_{i+1}(u)$ where $u$ is the grandparent of $e$ in $T_i$. This constant matrix in turn affects exactly one constant matrix in the next higher level, and so on. Since the effect at each level can be computed in $O(k^3)$ time (due to matrix multiplication) and there are $O(\log N)$ levels of equations, the update can be accomplished in $O(k^3 \log N)$ time. The constant $k^3$ is actually pessimistic, because faster matrix multiplication algorithms exist.

The update procedure is given in Figure 5. Update is initially called as Update$(\lambda(E) = e, i)$ where $E$ is a leaf, $i$ the level at which it was raked, and $e$ is the new evidence. This operation will start a sequence of $O(\log N)$ calls to function $\lambda$-Update $(X = Term, i)$ as the change will propagate to $\log N$ equations.

*FUNCTION* $\lambda$-QUERY $(x)$

We look up the equation associated with $\lambda(x)$ in $T_{ind(x)}$.

Case 1: $x$ is a leaf. Then the equation is of the form: $\lambda(x) = e$ where $e$ is known. In this case we return $e$.

Case 2: The equation associated with $\lambda(x)$ is of the form

$$\lambda(x) \quad = \quad A_i(x) \cdot \lambda(y) * B_i(x) \cdot \lambda(z)$$

where $z$ is a leaf and therefore $\lambda(z)$ is known. In this case we return

$$A_i(X) \cdot \lambda\text{-QUERY } (y) * B_i(X) \cdot \lambda(z)$$

The case where $y$ is the leaf is analogous.

Figure 4: Function to compute the $\lambda$ value of a node.

### 3.4 $\pi$ Queries

It is relatively easy to use a similar recursive procedure to perform $\pi(x)$ queries. Unfortunately, this approach yields an $O(\log^2 N)$-time algorithm if we simply use recursion to calculate $\pi$ terms and calculate $\lambda$ terms using our earlier procedure. This is because there will be $O(\log N)$ recursive calls to calculate $\pi$ values, but each is defined by an equation that also involves a $\lambda$ term taking $O(\log N)$ time to compute.

To achieve $O(\log N)$ time, we shall instead implement $\pi(x)$ queries by defining a procedure CALC$\pi\lambda$ $(x, i)$ which returns a triple of vectors $\langle P, L, R\rangle$ such that $P = \pi(x)$, $L = \lambda(y)$ and $R = \lambda(z)$ where $y$ and $z$ are the left and right children, respectively, of $x$ in $T_i$.

To compute $\pi(x)$ for some node $x$ we can do the following. Let $i = ind(x)$. The equation for $\pi(x)$ in $T_i$ is of the form:

$$\pi(x) \quad = \quad D_i(x) \cdot (\pi(u) * C_i(x) \cdot \lambda(v))$$

where $u$ is the parent of $x$ in $T_i$ and $v$ its sibling. We then call procedure CALC$\pi\lambda$ $(u, i+1)$ which will return the triple $\langle \pi(u), \lambda(v), \lambda(x)\rangle$, from which we immediately can compute $\pi(x)$ using the above equation.

Procedure CALC$\pi\lambda$ $(x, i)$ can be implemented in the following fashion.

Case 1: If $T_i$ is a 3-node tree with $x$ as its root, then both children of $x$ are leaves, hence their $\lambda$ values are known, and $\pi(x)$ is a given sequence of prior probabilities for $x$.

Case 2: If $x$ does not appear in $T_{i+1}$, then one of $x$'s children is a leaf, say $e$ which is raked at level $i$. Let $z$ be the other child. We call CALC$\pi\lambda$ $(u, i+1)$, where $u$ is the parent of $x$ in $T_i$, and receive back $\langle \pi(u), \lambda(z), \lambda(v)\rangle$ or $\langle \pi(u), \lambda(v), \lambda(z)\rangle$ according to whether $x$

*FUNCTION* $\lambda$-Update (*Term* = *Value*, $i$)

1. Find the (at most one) equation in $T_i$, defining some $A_i$ or $B_i$, in which *Term* appears on the right-hand side; let *Term'* be the matrix defined by this equation (*i.e.*, its left-hand side).
2. Update *Term'*; let *Value* be the new value.
3. Call $\lambda$-Update (*Term'* = *Value*, $i+1$) recursively.

Figure 5: The update procedure.

was a left or right child of $u$ in $T_i$ (and $v$ is $u$'s other child). We can now compute $\pi(x)$ from $\pi(u)$ and $\lambda(v)$, and we have $\lambda(e)$ and $\lambda(z)$, so we can return the necessary triple.

Specifically,

$$\pi(x) = \begin{cases} D_i(x) \cdot (\pi(u) * A_{i+1}(u) \cdot \lambda(v)) \\ D_i(x) \cdot (\pi(u) * B_{i+1}(u) \cdot \lambda(v)) \end{cases}$$

where the choice depends on whether $x$ is the right or left child, respectively, of $u$ in $T_i$.

Case 3: If $x$ does appear in $T_{i+1}$, then we call Calc$\pi\lambda$ $(x, i+1)$. This returns the correct value of $\pi(x)$. For any child $z$ of $x$ in $T_i$ that remains a child of $x$ in $T_{i+1}$, it also returns the correct value of $\lambda(z)$. If $z$ is a child of $x$ that does not occur in $T_{i+1}$, then it must be the case that $z$ was raked at level $i$ so that one of $z$'s children, say $e$, is a leaf and let the other child be $q$. In this situation Calc$\pi\lambda$ $(x, i+1)$ has returned the value of $\lambda(q)$ and we can compute

$$\lambda(z) = A_i(z) \cdot \lambda(e) * B_i(z) \cdot \lambda(q)$$

and return this value.

In all three cases, there is a constant amount of work done in addition to a single recursive call that uses equations at a higher level. Since there are $O(\log N)$ levels of equations, each requiring only matrix by vector multiplication, the total work done is $O(k^2 \log N)$.

## 4. Extended Example

In this section we illustrate the application of our algorithms to a specific example. Consider the sequence of contracted trees shown in Figure 6. Corresponding to these trees we have

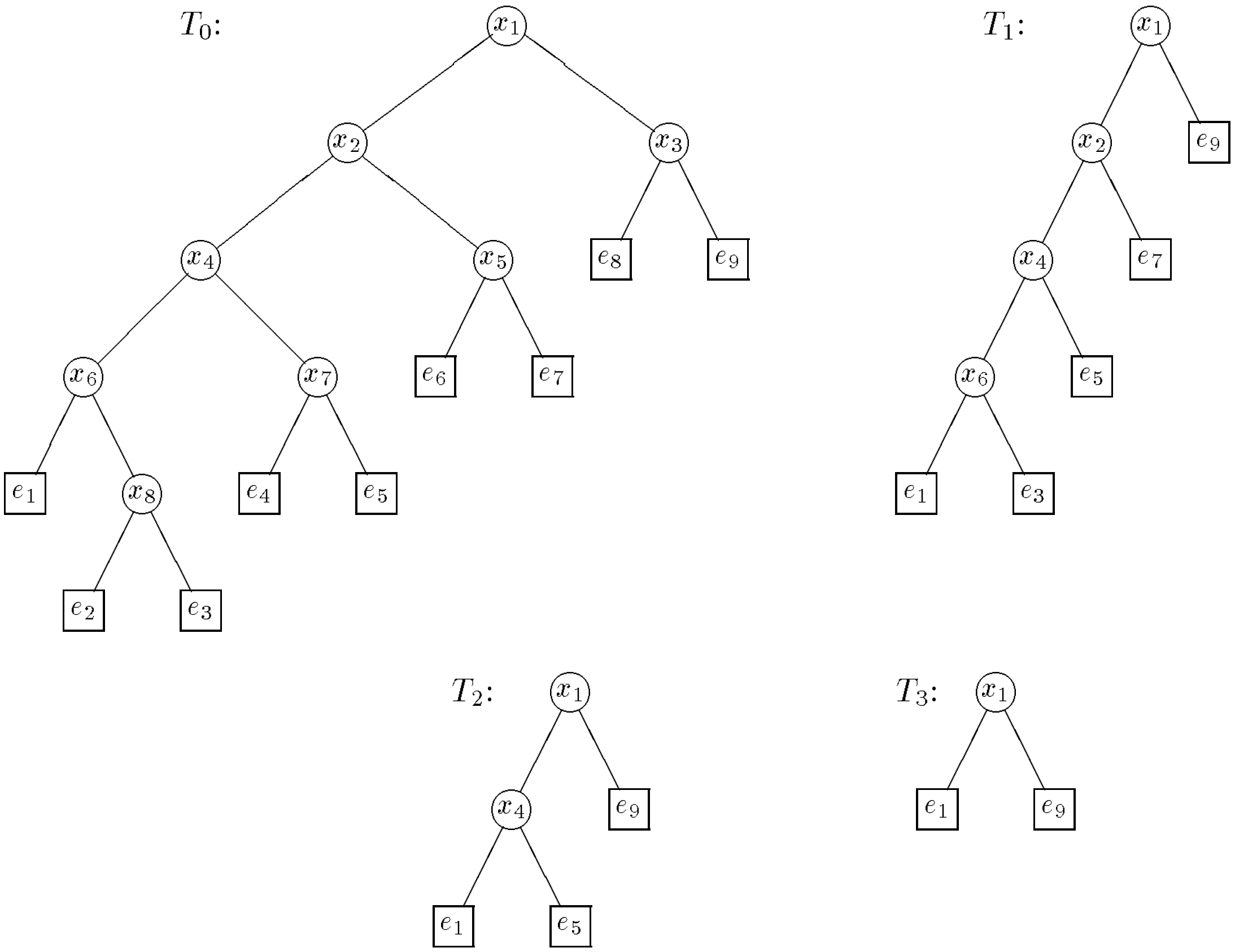

Figure 6: Example of tree contraction.

such equations as the following:

For $T_0$:

$$\begin{array}{rclrcl} \lambda(x_1) & = & A_0(x_1)\cdot\lambda(x_2) * B_0(x_1)\cdot\lambda(x_3) & \pi(x_2) & = & D_0(x_2)\cdot(\pi(x_1) * C_0(x_2)\cdot\lambda(x_3)) \\ & \vdots & & & \vdots & \end{array}$$

For $T_1$:

$$\begin{array}{rclrcl} \lambda(x_1) & = & A_1(x_1)\cdot\lambda(x_2) * B_1(x_1)\cdot\lambda(e_9) & \pi(x_2) & = & D_1(x_2)\cdot(\pi(x_1) * C_1(x_2)\cdot\lambda(e_9)) \\ & \vdots & & & \vdots & \end{array}$$

For $T_2$:

$$\begin{array}{rclrcl} \lambda(x_1) & = & A_2(x_1)\cdot\lambda(x_4) * B_2(x_1)\cdot\lambda(e_9) & \pi(x_4) & = & D_2(x_4)\cdot(\pi(x_1) * C_2(x_4)\cdot\lambda(e_9)) \\ & \vdots & & & \vdots & \end{array}$$

For $T_3$:

$$\lambda(x_1) = A_3(x_1)\cdot\lambda(e_1) * B_3(x_1)\cdot\lambda(e_9)$$

Now consider, for instance, the effect of an update for $e_2$. Since it is raked immediately, the new value of $\lambda(e_2)$ is incorporated in:

$$B_1(x_6) = B_0(x_6)\cdot \mathrm{Diag}_{A_0(x_8)\cdot\lambda(e_2)}\cdot B_0(x_8)$$

From subsequent Rake operations we know that $A_2(x_4)$ depends on $B_1(x_6)$, and $A_3(x_1)$ depends on $A_2(x_4)$, so we must also update these values as follows:

$$\begin{array}{rcl} A_2(x_4) & = & A_1(x_4)\cdot \mathrm{Diag}_{B_1(x_6)\cdot\lambda(e_3)}\cdot A_1(x_6) \\ A_3(x_1) & = & A_2(x_1)\cdot \mathrm{Diag}_{B_2(x_4)\cdot\lambda(e_5)}\cdot A_2(x_4) \end{array}$$

Finally, consider a query for $x_7$. Since $x_7$ is raked together with $e_5$ in $T_0$, we follow the steps outlined above and generate the following calls: Calc$\pi\lambda\,(x_7, 0)$, Calc$\pi\lambda\,(x_4, 1)$, Calc$\pi\lambda\,(x_4, 2)$, and Calc$\pi\lambda\,(x_1, 3)$. This provides us with $\pi(x_7)$. In this case, $\lambda(x_7)$ is particularly easy to compute since both $x_7$'s children are leaf nodes. Then we simply compute $\pi(x_7) * \lambda(x_7)$ and then normalize, giving us the conditional marginal distribution $Bel(x_7)$ as required.

## 5. Join Trees

Perhaps the best-known technique for computing with arbitrary (*i.e.*, *not* singly-connected) Bayesian networks uses the idea of *join trees* (junction trees) (Lauritzen & Spiegelhalter, 1988). In many ways a join tree can be thought of as a causal tree, albeit one with somewhat special structure. Thus the algorithm in the previous section can be applied. However, the structure of a join tree permits some optimization, which we describe in this section. This becomes especially relevant in the next section, where we use the join-tree technique to show how $O(\log N)$ updates and queries can be done for arbitrary polytrees. Our review of join-trees and their utility is extremely brief and quite incomplete; for clear expositions see, for instance, Spiegelhalter et al. (1993) and Pearl (1988).

Given any Bayesian network, the first step towards constructing a join-tree is to *moralize* the network: insert edges between every pair of parents of a common node, and then treat all edges in the graph as being undirected (Spiegelhalter et al., 1993). The resulting undirected graph is called the *moral* graph. We are interested in undirected graphs that are *chordal*: every cycle of length 4 or more should contain a chord (*i.e.*, an edge between two nodes that are non-adjacent in the cycle). If the moral graph is not chordal, it is necessary to add edges to make it so; various techniques for this *triangulation* stage are known (for instance, see Spiegelhalter et al., 1993).

If $p$ is a probability distribution represented in a Bayesian network $G = (V, E)$, and $M = (V, F)$ is the result of moralizing and then triangulating $G$, then:

1. $M$ has at most $|V|$ cliques,[4] say $C_1, \ldots, C_{|V|}$.
2. The cliques can be ordered so that for each $i > 1$ there is some $j(i) < i$ such that
$$C_i \cap C_{j(i)} = C_i \cap (C_1 \cup C_2 \cup \ldots \cup C_{i-1}.)$$
The tree $T$ formed by treating the cliques as nodes, and connecting each node $C_i$ to its "parent" $C_{j(i)}$, is called a *join tree.*
3. $p = \prod_i p(C_i | C_{j(i)})$
4. $p(C_i | C_{j(i)}) = p(C_i | C_{j(i)} \cap C_i)$

From 2 and 3, we see that if we direct the edges in $T$ away from the "parent" cliques, the resulting directed tree is in fact a Bayesian causal tree that can represent the original distribution $p$. This is true no matter what the form of the original graph. Of course, the price is that the cliques may be large, and so the domain size (the number of possible values of a clique node) can be of exponential size. This is why this technique is not guaranteed to be efficient.

We can use the Rake technique of Section 2 on the directed join tree without any modification. However, property 4 above shows that the conditional probability matrices in the join tree have a special structure. We can use this to gain some efficiency. In the following, let $k$ be the domain size of the variables in $G$ as usual. Let $n$ be the maximum size of cliques in the join tree; without loss of generality we can assume that all cliques are of the same size (because we can add "dummy" variables). Thus the domain size of each clique is $K = k^n$. Finally, let $c$ be the maximum intersection size of a clique and its parent (*i.e.*, $|C_{j(i)} \cap C_i|$) and $L = k^c$.

In the standard algorithm, we would represent $p(C_i | C_{j(i)})$ as a $K \times K$ matrix, $M_{C_i|C_{j(i)}}$. However, $p(C_i | C_{j(i)} \cap C_i)$ can be represented as a smaller $L \times K$ matrix, $M_{C_i|C_{j(i)} \cap C_i}$. By property 4 above, $M_{C_i|C_{j(i)}}$ is identical to $M_{C_i|C_{j(i)} \cap C_i}$, except that many rows are repeated. Thus there is a $K \times L$ matrix $J$ such that

$$M_{C_i|C_{j(i)}} = J \cdot M_{C_i|C_{j(i)} \cap C_i}.$$

($J$ is actually a simple matrix whose entries are 0 and 1, with exactly one 1 per row; however we do not use this fact.)

4. A *clique* is a maximal completely-connected subgraph.

Our claim is that, in the case of join trees, the following is true. First, the matrices $A_i$ and $B_i$ used in the Rake algorithm can be stored in factored form, as the product of two matrices of dimension $K \times L$ and $L \times K$ respectively. So, for instance, we factor $A_i$ as $A_i^l \cdot A_i^r$. We never need to explicitly compute, or store, the full matrices. As we have just seen, this claim is true when $i = 0$ because the $M$ matrices factor this way. The proof for $i > 1$ uses an inductive argument, which we illustrate below. The second claim is that, when the matrices are stored in factored form, all the matrix multiplications used in the Rake algorithm are of one of the following types: 1) an $L \times K$ matrix times a $K \times L$ matrix, 2) an $L \times K$ matrix times a $K \times K$ diagonal matrix, 3) an $L \times L$ matrix times an $L \times K$ matrix, or 4) an $L \times K$ matrix times a vector.

To prove these claims consider, for instance, the equation defining $B_{i+1}$ in terms of lower-level matrices. From Section 2, $B_{i+1}(u) = B_i(u) \cdot \mathrm{Diag}_{A_i(x) \cdot \lambda(e)} \cdot B_i(x)$. But, by assumption, this is:

$$(B_i^l(u) \cdot B_i^r(u)) \cdot \mathrm{Diag}_{(A_i^l(x) \cdot A_i^r(x)) \cdot \lambda(e)} \cdot (B_i^l(x) \cdot B_i^l(x)),$$

which, using associativity, is clearly equivalent to

$$B_i^l(u) \cdot \left[ ((B_i^r(u) \cdot \mathrm{Diag}_{A_i^l(x) \cdot (A_i^r(x) \cdot \lambda(e))}) \cdot B_i^l(x)) \cdot B_i^l(x) \right].$$

However, every multiplication in this expression is one of the forms stated earlier. Identifying $B_{i+1}^l(u)$ as $B_i^l(u)$ and $B_{i+1}^r(u)$ as the bracketed part of the expression proves this case, and of course the case where we rake a left child (so that $A_{i+1}(u)$ is updated) is analogous. Thus, even using the most straightforward technique for matrix multiplication, the cost of updating $B_{i+1}$ is $O(KL^2) = O(k^{n+2c})$. This contrasts with $O(K^3)$ if we do not factor the matrices, and may represent a worthwhile speedup if $c$ is small. Note that the overall time for an update using this scheme is $O(k^{n+2c} \log N)$. Queries, which only involve matrix by vector multiplication, require $O(k^{n+c} \log N)$ time.

For many join trees the difference between $N$ and $\log N$ is unimportant, because the clique domain size $K$ is often enormous and dominates the complexity. Indeed, $K$ and $L$ may be so large that we cannot represent the required matrices explicitly. Of course, in such cases our technique has little to offer. But there will be other cases in which the benefits will be worthwhile. The most important general class in which this is so, and our immediate reason for presenting the technique for join trees, is the case of polytrees.

## 6. Polytrees

A polytree is a singly connected Bayesian network; we drop the assumption of Section 2 that each node has at most one parent. Polytrees offer much more flexibility than causal trees, and yet there is a well-known process that can update and query in $O(N)$ time, just as for causal trees. For this reason polytrees are an extremely popular class of networks.

We suspect that it is possible to present an $O(\log N)$ algorithm for updates and queries in polytrees, as a direct extension of the ideas in Section 2. Instead we propose a different technique, which involves converting a polytree to its join tree and then using the ideas of the preceding section. The basis for this is the simple observation that the join tree of a polytree is already chordal. Thus (as we show in detail below) little is lost by considering the join tree instead of the original polytree. The specific property of polytrees that we require is the following. We omit the proof of this well-known proposition.

**Proposition 4:** If $T$ is the moral graph of a polytree $P = (V, E)$ then $T$ is chordal, and the set of maximal cliques in $T$ is $\{\{v\} \cup \mathit{parents}(v) : v \in V\}$.

Let $p$ be the maximum number of parents of any node. From the proposition, every maximal clique in the join tree has at most $p+1$ variables, and so the domain size of a node in the join tree is $K = k^{p+1}$. This may be large, but recall that the conditional probability matrix in the original polytree, for a variable with $p$ parents, has $K$ entries anyway since we must give the conditional distribution for every combination of the node's parents. Thus $K$ is really a measure of the size of the polytree itself.

It now follows from the proposition above that we can perform query and update in polytrees in time $O(K^3 \log N)$, simply by using the algorithm of Section 2 on the directed join tree. But, as noted in Section 5, we can do better. Recall that the savings depend on $c$, the maximum size of the intersection between any node and its parent in the join tree. However, when the join tree is formed from a polytree, no two cliques can share more than a single node. This follows immediately from Proposition 4, for if two cliques have more than one node in common then there must be either two nodes that share more than one parent, or else a node and one of its parents that both share yet another parent. Neither of these is consistent with the network being a polytree. Thus in the complexity bounds of Section 5, we can put $c = 1$. It follows that we can process updates in $O(Kk^{2c} \log N) = O(k^{p+3} \log N)$ time and queries in $O(k^{p+2} \log N)$.

## 7. Application: Towards Automated Site-Specific Muta-Genesis

An experiment which is commonly performed in biology laboratories is a procedure where a particular site in a protein is changed (*i.e.*, a single amino-acid is mutated) and then tested to see whether the protein settles into a different conformation. In many cases, with overwhelming probability the protein does not change its secondary structure outside the mutated region. This process is often called muta-genesis. Delcher et al. (1993) developed a probabilistic model of a protein structure which is basically a long chain. The length of the chain varies between 300–500 nodes. The nodes in the network are either protein-structure nodes (*PS*-nodes) or evidence nodes (*E*-nodes). Each *PS*-node in the network is a discrete random variable $X_i$ that assumes values corresponding to descriptors of secondary sequence structure: helix, sheet or coil. With each *PS*-node the model associates an evidence node that corresponds to an occurrence of a particular subsequence of amino acids at a particular location in the protein.

In our model, protein-structure nodes are finite strings over the alphabet $\{h, e, c\}$. For example the string *hhhhhh* is a string of six residues in an $\alpha$-helical conformation, while *eecc* is a string of two residues in a $\beta$-sheet conformation followed by two residues folded as a coil. Evidence nodes are nodes that contain information about a particular region of the protein. Thus, the main idea is to represent physical and statistical rules in the form of a probabilistic network.

In our first set of experiments we converged on the following model that, while clearly biologically naive, seems to match in prediction accuracy many existing approaches such as neural networks. The network looks like a set of *PS*-nodes connected as a chain. To each such node we connect a single evidence node. In our experiments the *PS*-nodes are strings of length two or three over the alphabet $\{h, e, c\}$ and the evidence nodes are strings of the same length over the set of amino acids. The following example clarifies our representation. Assume we have a string of amino acids *GSAT*. We model the string as a network comprised of three evidence nodes *GS*, *SA*, *AT* and three *PS*-nodes. The network is shown in Figure 7. A correct prediction will assign the values *cc*, *ch*, and *hh* to the *PS*-nodes as shown in the figure.

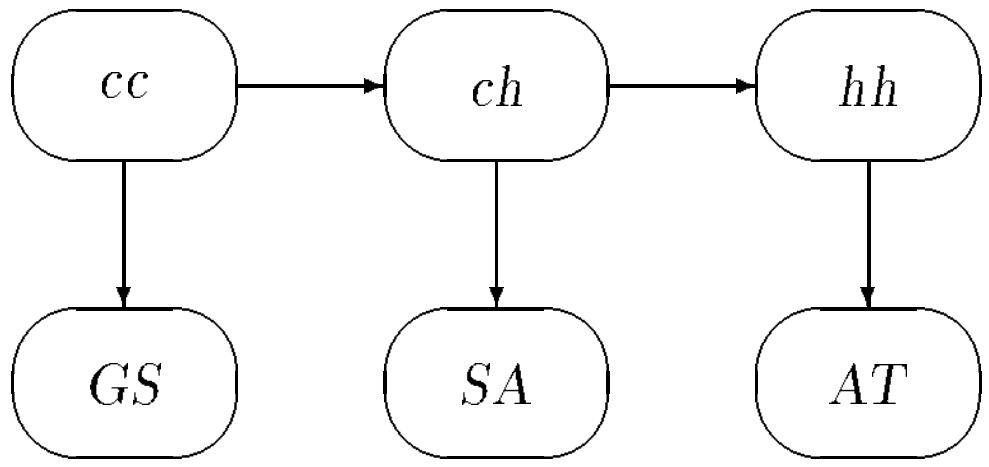

Figure 7: Example of causal tree model using pairs, showing protein segment *GSAT* with corresponding secondary structure *cchh*.

Now that we have a probabilistic model, we can test the robustness of the protein or whether small changes in the protein affect the structure of certain critical sites in the protein. In our experiments, the probabilistic network performs a "simulated evolution" of the protein, namely the simulator repeatedly mutates a region in the chain and then tests whether some designated sites in the protein that are coiled into a helix are predicted to remain in this conformation. The main goal of the experiment was to test if stable bonds far away from the mutated location were affected. Our previous results (Delcher et al., 1993) support the current thesis in the biology community, namely that local distant changes rarely affect structure.

The algorithms we presented in the previous sections of the paper are perfectly suited for this type of application and are predicted to generate a factor of 10 improvement in efficiency over the current brute-force implementation presented by Delcher et al. (1993) where each change is propagated throughout the network.

## 8. Summary

This paper has proposed several new algorithms that yield a substantial improvement in the performance of probabilistic networks in the form of causal trees. Our updating procedures absorb sufficient information in the tree such that our query procedure can compute the correct probability distribution of any node given the current evidence. In addition, all procedures execute in time $O(\log N)$, where $N$ is the size of the network. Our algorithms are expected to generate orders-of-magnitude speed-ups for causal trees that contain long paths (not necessarily chains) and for which the matrices of conditional probabilities are relatively small. We are currently experimenting with our approach with singly connected networks (polytrees). It is likely to be more difficult to generalize the techniques to general networks. Since it is known that the general problem of inference in probabilistic networks is $\mathcal{NP}$-hard (Cooper, 1990), it obviously is not possible to obtain polynomial-time incremental solutions of the type discussed in this paper for general probabilistic networks. The other natural open question is extending the approach developed in this paper to other dynamic operations on probabilistic networks such as addition and deletion of nodes and modifying the matrices of conditional probabilities (as a result of learning).

It would also be interesting to investigate the practical logarithmic-time parallel algorithms for probabilistic networks on realistic parallel models of computation. One of the main goals of massively parallel AI research is to produce networks that perform real-time inference over large knowledge-bases very efficiently (*i.e.*, in time proportional to the depth of the network rather than the size of the network) by exploiting massive parallelism. Jerry Feldman pioneered this philosophy in the context of neural architectures (see Stanfill and Waltz, 1986, Shastri, 1993, and Feldman and Ballard, 1982). To achieve this type of performance in the neural network framework, we typically postulate a parallel hardware that associates a processor with each node in a network and typically ignores communication requirements. With careful mapping to parallel architectures one can indeed achieve efficient parallel execution of specific classes of inference operations (see Mani and Shastri, 1994, Kasif, 1990, and Kasif and Delcher, 1992). The techniques outlined in this paper presented an alternative architecture that supports very fast (sub-linear time) response capability on sequential machines based on preprocessing. However, our approach is obviously limited to applications where the number of updates and queries at any time is constant. One would naturally hope to develop parallel computers that support real-time probabilistic reasoning for general networks.

## Acknowledgements

Simon Kasif's research at Johns Hopkins University was sponsored in part by National Science foundation under Grants No. IRI-9116843, IRI-9223591 and IRI-9220960.

## References

Berger, T., & Ye, Z. (1990). Entropic aspects of random fields on trees. *IEEE Trans. on Information Theory*, *36*(5), 1006–1018.

Chelberg, D. M. (1990). Uncertainty in interpretation of range imagery. In *Proc. Intern. Conference on Computer Vision*, pp. 654–657.

Cohen, R. F., & Tamassia, R. (1991). Dynamic trees and their applications. In *Proceedings of the 2nd ACM-SIAM Symposium on Discrete Algorithms*, pp. 52–61.

Cooper, G. (1990). The computational complexity of probabilistic inference using bayes belief networks. *Artificial Intelligence*, *42*, 393–405.

Delcher, A., & Kasif, S. (1992). Improved decision making in game trees: Recovering from pathology. In *Proceedings of the 1992 National Conference on Artificial Intelligence.*

Delcher, A. L., Kasif, S., Goldberg, H. R., & Hsu, B. (1993). Probabilistic prediction of protein secondary structure using causal networks. In *Proceedings of 1993 International Conference on Intelligent Systems for Computational Biology*, pp. 316–321.

Duda, R., & Hart, P. (1973). *Pattern Classification and Scene Analysis*. Wiley, New York.

Feldman, J. A., & Ballard, D. (1982). Connectionist models and their properties. *Cognitive Science*, *6*, 205–254.

Frederickson, G. N. (1993). A data structure for dynamically maintaining rooted trees. In *Proc. 4th Annual Symposium on Discrete Algorithms*, pp. 175–184.

Hel-Or, Y., & Werman, M. (1992). Absolute orientation from uncertain data: A unified approach. In *Proc. Intern. Conference on Computer Vision and Pattern Recognition*, pp. 77–82.

Karp, R. M., & Ramachandran, V. (1990). Parallel algorithms for shared-memory machines. In Van Leeuwen, J. (Ed.), *Handbook of Theoretical Computer Science*, pp. 869–941. North-Holland.

Kasif, S. (1990). On the parallel complexity of discrete relaxation in constraint networks. *Artificial Intelligence*, *45*, 275–286.

Kasif, S., & Delcher, A. (1994). Analysis of local consistency in parallel constraint networks. *Artificial Intelligence*, *69*.

Kosaraju, S. R., & Delcher, A. L. (1988). Optimal parallel evaluation of tree-structured computations by raking. In Reif, J. H. (Ed.), *VLSI Algorithms and Architectures: Proceedings of 1988 Aegean Workshop on Computing*, pp. 101–110. Springer Verlag. LNCS 319.

Lauritzen, S., & Spiegelhalter, D. (1988). Local computations with probabilities on graphical structures and their applications to expert systems. *J. Royal Statistical Soc. Ser. B*, *50*, 157–224.

Mani, D., & Shastri, L. (1994). Massively parallel reasoning with very large knowledge bases. Tech. rep., Intern. Computer Science Institute.

Miller, G. L., & Reif, J. (1985). Parallel tree contraction and its application. In *Proceedings of the 26th IEEE Symposium on Foundations of Computer Science*, pp. 478–489.

Pearl, J. (1988). *Probabilistic Reasoning in Intelligent Systems*. Morgan Kaufmann.

Peot, M. A., & Shachter, R. D. (1991). Fusion and propagation with multiple observations in belief networks. *Artificial Intelligence*, *48*, 299–318.

Rachlin, J., Kasif, S., Salzberg, S., & Aha, D. (1994). Towards a better understanding of memory-based and bayesian classifiers. In *Proceedings of the Eleventh International Conference on Machine Learning*, pp. 242–250 New Brunswick, NJ.

Shastri, L. (1993). A computational model of tractable reasoning: Taking inspiration from cognition. In *Proceeding of the 1993 Intern. Joint Conference on Artificial Intelligence*. AAAI.

Spiegelhalter, D., Dawid, A., Lauritzen, S., & Cowell, R. (1993). Bayesian analysis in expert systems. *Statistical Science*, *8*(3), 219–283.

Stanfill, C., & Waltz, D. (1986). Toward memory-based reasoning. *Communications of the ACM*, *29*(12), 1213–1228.

Wilsky, A. (1993). Multiscale representation of markov random fields. *IEEE Trans. Signal Processing*, *41*, 3377–3395.